\documentclass{article}
\usepackage{pslatex}
\usepackage{graphics}
\usepackage{rotating}
\usepackage[table]{xcolor}
\usepackage{caption}
\usepackage{hyperref}
\usepackage{amsthm}
\usepackage{amsmath}
\usepackage{enumitem}
\usepackage{amssymb}
\theoremstyle{plain}
\usepackage{tabularx}
\usepackage{multirow}

\begin{document}

\title{A Physical and Mathematical Framework for the Semantic Theory of Evolution}

\author{Guido Fioretti\\University of Bologna\\Management Department}
\maketitle

\begin{abstract}
The Semantic Theory of Evolution (STE) takes the existence of a number of arbitrary communication codes as a fundamental feature of life, from the genetic code to human cultural communication codes. Their arbitrariness enables, at each level, the selection of one out of several possible correspondences along with the generation of meaning. STE enables more novelties to emerge and suggests a greater variety of potential life forms.

With this paper I ground STE on physical theories of meaningful information. Furthermore,  I show that key features of the arbitrary communication codes employed by living organisms can be expressed by means of Evidence Theory (ET).

In particular, I adapt ET to organisms that merely react to sequences of stimuli, explain its basics for organisms that are capable of prediction, and illustrate an unconventional version suitable  for the most intricate communication codes employed by humans. Finally, I  express the natural trend towards ambiguity reduction in  terms of information entropy minimization along with thermodynamic entropy maximization.
\end{abstract}


\newpage

\section{Introduction}     \label{sec:intro}

The \emph{Semantic Theory of Evolution} (STE) \cite{barbieri-85} maintains that generation of meaning is an essential feature of life. In particular, STE stresses that one distinguishing feature of life is the arbitrariness of its communication codes, starting with the genetic code and moving up to the histone code, the sugar code, the cytoskeleton code and, finally, the cultural communication codes so typical of man \cite{barbieri-18BS} \cite{kun-21BS} \cite{prinz-24BT} \cite{CB}. These codes are arbitrary in the sense that they connect two domains in some way, and yet many other ways would be equally possible; for instance, the genetic code makes certain triplets of nucleotides (codons) correspond to specific amino-acids, but evolution could have selected other correspondences  \cite{barbieri-85} \cite{barbieri-24}.

In particular, in the early stages of life the genetic code is likely to have been ambiguous in the sense that one and the same codon would randomly command one out of several amino-acids, which would synthetize so-called statistical proteins  in their turn \cite{woese-65PNAS}. In other words, in the ancestral, ambiguous genetic code, one single codon could have several meanings. However, evolutionary pressures eventually favoured the selection of one specific amino-acid for each codon, generating the redundant but non-ambiguous genetic code that characterizes life today \cite{barbieri-85} \cite{barbieri-24}. 

Similar reasonings can be applied to all other codes that have appeared after the genetic code in the history of life. All of them have provided and still provide opportunities to generate novel correspondences in a space enabled by their arbitrariness, and all of them are subject to more or less successful evolutionary pressures to decrease their  ambiguity. Indeed, STE points to the existence of a hierarchy of arbitrary communication codes, from the genetic code to the arbitrary codes for infra- and inter-cellular messages, to animal communication and human cultural codes \cite{kun-21BS} \cite{prinz-22BS} \cite{prinz-23BS}.

In principle, STE can be combined with other theories  in order to suggest the existence of one and the same organizing principle for all  living beings, from unicellular organisms to animal and human organizations and societies. For instance, \emph{biosemiotics} derives agency from the generation of meaning  \cite{sharov-tonnessen-21}, or, Eigen-Schuster's \emph{hypercycles} can generate hierarchical structures ---  such as the syntax of human languages, as well as Science  ---  insofar they receive meaningful rewards \cite{marchetti-83TFSC}.

Viewing life as based on arbitrary communication codes has a substantial impact on its capability of generating novelties. While the Neo-Darwinian Synthesis (NDS) ascribes the origin of novelty exclusively to the random mutation of genes, STE adds the many more possibilities that arise from selecting one out of many arbitrary correspondences in communication channels at all levels. In a nutshell, while NDS is based on chance acting on copying information, STE adds the possibility for chance to act on coding information as well \cite{barbieri-03} \cite{barbieri-24}. Novelty is not limited to the generation of novel elements, but arises from the generation of novel correspondences between  elements as well and, just like in a graph the number of potential links between nodes grows roughly with the square of the number of nodes, STE suggests the possibility space is explored at a much higher rate than NDS allows.

This difference reverberates, among else, onto the relation between physics and life sciences. While NDS views life as a highly improbable meeting of the right components at the right time and the right place in a primordial soup, STE rather views life as a very natural outcome of information transmission paths enabled by energy sources and coupled to environmental rewards. Thus, while NDS understands life as a very unlikely exception to the Second Law of Thermodynamics \cite{schrodinger-92}, STE rather views it as high-probability processes induced by the structures of the possibility spaces where the Second Law of Thermodynamics unfolds \cite{jeffery-pollack-rovelli-19E} \cite{jeffery-rovelli-20TN}.

In general, STE is  presented by embedding concepts from semiotics within theoretical biology \cite{barbieri-15}. By contrast, this article combines STE with recent insights from physics, namely the formalization of \emph{meaningful information} \cite{rovelli-18III} \cite{kolchinsky-wolpert-18IF}, as well as an extension of the mathematics of information communication and evaluation  according to \emph{Evidence Theory} (ET) \cite{shafer-76}. On the one hand, the physical theory of meaningful information resolves certain  disputes within STE. On the other hand, ET enables formalization of arbitrariness reduction for several classes of living beings.

The rest of this article is organized as follows. The ensuing \S~\ref{sec:theory} illustrates the physical concept of  meaningful information and the basics of ET in \S~\ref{subsec:Meaning} and \S~\ref{subsec:ET}, respectively. Subsequently, \S~\ref{sec:three} adapts ET to three classes of living beings, namely unicellular organisms, plants and animals that are incapable of prediction in \S~\ref{subsec:genetic}, animals with a nervous system sufficiently developed to figure out future states in \S~\ref{subsec:wolf-snake} and primates who are capable of figuring out what others are thinking about them in~\S~\ref{subsec:ToM}, respectively. Subsequently, \S~\ref{sec:entropy} casts evolutionary pressures on information transmission in terms of information maximization. Finally,  \S~\ref{sec:conclusions} recapitulates the previous constructs tracing parallels with  the logic of deduction, induction and abduction, respectively.

\section{Preliminary Concepts}   \label{sec:theory}

In this section I expound the basics of the physical theory of meaningful information and the general framework of ET, respectively. Henceforth, ET will be assumed to employ different rules to combine evidence, which will be exponded in \S~\ref{subsec:genetic}, \S~\ref{subsec:wolf-snake} and \S~\ref{subsec:ToM}, respectively.

\subsection{Meaning and Interpretation}  \label{subsec:Meaning}

Shannon's \emph{Information Theory} (IT) concerns the transmission of information carried by dicrete signals through a noisy channel. Signals correspond to a set of characters, which can be emitted by the source with a probability distribution known to the receiver. In this context, \emph{information} is the reduction of uncertainty for the receiver upon receiving a character. Thus, information is  highest when a character is received, that had the lowest probability to be emitted. The average of the information obtained by receiving each possible character is called \emph{Information Entropy}. Information entropy has the same functional form as thermodynamic entropy and it is maximum when characters are equiprobable \cite{shannon-weaver-49}.

Shannon defined information in a way that makes it independent of meaning. Indeed, the above definition of information is ultimately a measure of correlation between emitter and receiver.

Recently, a physical definition of \emph{meaningful information} has been proposed (also known as \emph{semantic information} by contrast to Shannon's \emph{synctactic information}), based on evolutionary theory. In a nutshell, a portion of Shannon's information that living organisms exchange with their environment is \emph{meaningful} to them insofar it increases their chance of survival \cite{rovelli-18III} \cite{kolchinsky-wolpert-18IF}. For instance, the location of nutrients or poisonous substances is meaningful for bacteria, whereas the colour of the surface on which they rest is generally irrelevant for them.

Henceforth, I shall imply that whenever a piece of information owned by a living organism receives a feedback from the environment, this piece of information acquires a \emph{meaning} and I shall use the term \emph{interpretation} for the process of receiving a meaning. This process is purely mechanical, in the sense that it does not require anything like ``intentionality''  or other capabilities that cannot be ascribed to unicellular organisms. In the aforedescribed sense, interpretation can be as simple as an environmental feedback being coupled to one (or a set of, or a combination of) nucleic acid(s) out of pure chance, this coupling being eventually reinforced through repetition.

Semiotics has quite a  different understanding of what contitutes ``meaning'' and ``interpretation.'' Semiotics studies \emph{signs} that communicate a \emph{meaning} to their \emph{interpreter}. This framework is evidently derived from human communication, and it can be easily extended to include animal communication as well \cite{sebeok-10VI}. However, stretching it to living beings that do not even have a nervous system is problematic, because re-naming a relatively simple component as an ``interpreter'' is dangerously close to vitalism, or some other form of magic. This dilemma generated a still unsettled debate, where  the concept of interpretation is either downgraded when it must be applied to unicellular organisms, or outright rejected \cite{kull-20CF} \cite{barbieri-24}.

The physical notion of meaningful information can resolve this controversy, because interpretation at such basic levels as the genetic code either takes place out of random coupling of exogenous feed-backs to codes, without any interpreter, or with the help of such simple an ``interpreter'' that neither \textit{homunculi} nor elaborate nervous systems are needed. From this point of view, the elaborate concepts employed by semiotics appear to be derived from more basic physical and biological principles, and are only applicable to animals whose cognitive abilities are sufficiently sophisticated. Notably, a similar cultural operation has been put forward by employing the concept of ``consequences for the carrier'' which is equivalent to that of environmental feed-backs, albeit it has not been been linked to the notion of evolutionary fitness \cite{prinz-24B}.

\subsection{A Mathematical Theory of Evidence}   \label{subsec:ET}

ET \cite{shafer-76}, also known as ``Dempster-Shafer Theory'' or ``Belief Functions Theory,'' is a mathematical theory of uncertain reasoning that takes as  prototypical situation a judge evaluating testimonies, or a detective examining cues, rather than a gambler playing dice \cite{shafer-81S} \cite{shafer-tversky-85CS}.  This marks a sharp difference with   Probability Theory (PT) in at least two respects:

\begin{enumerate}
\item While gamblers playing dice know the set of possibilities (the six faces of a die), judges listening to testimonies and detectives looking for cues know that novel possibilities may appear. Unexpected denouements are, indeed, the salt of detective stories.

\item While gamblers face a set of disjoint possibilities (either face 1, or face 2, and so on), judges listening to testimonies and detectives looking for cues must combine coherent portions  of the information that they receive, while discarding the rest as irrelevant. This finds applications, among else, in the field of data fusion.
\end{enumerate}

Property (1) is essential in order to express the ability of living organisms to generate novelties by means of random mutations, which is further magnified by establishing conventions that are enabled by the arbitrariness of  communication codes. Property (2) is relevant because the coherence or incoherence of portions of information can be decisive insofar it concerns which meaning is assigned to them. In other words, arbitrariness can make communication code ambiguous to some extent, in which case coherence may shift the balance between alternative interpretations.

Correspondingly, ET makes assumptions that are different from those of PT. Specifically, the possibility set is called \emph{Frame of Discernment} (FoD) and translates the above properties as follows:~\footnote{There exist variants of ET which, in order to stay closer to PT, assimilate the FoD to a $\sigma$-algebra. These variants are ignored in the present work.}

\begin{enumerate}
\item The FoD is not a $\sigma$-algebra with respect to complementation. This ensures that uncertainty about novel possibilities cannot be hidden by defining a residual event subsuming any unexpected novelpossibility.

\item Possibilities appear in the FoD as sets that may be disjoint, or intersect, or being included in one another, or in the limit coincide. Disjoint sets ($A_i \cap A_j = \emptyset$) represent contradictory possibilities, intersecting sets ($A_i \cap A_j \neq \emptyset$, $A_i \nsubseteq  A_j$) represent partially coherent possibilities, whereas a set $A_i$ included or coinciding with $A_j$ ($A_i \subseteq A_j$) means that $A_i$ is coherent with $A_j$.
\end{enumerate}

\begin{figure}
\center
\fbox{\resizebox*{0.8\textwidth}{!}{\includegraphics{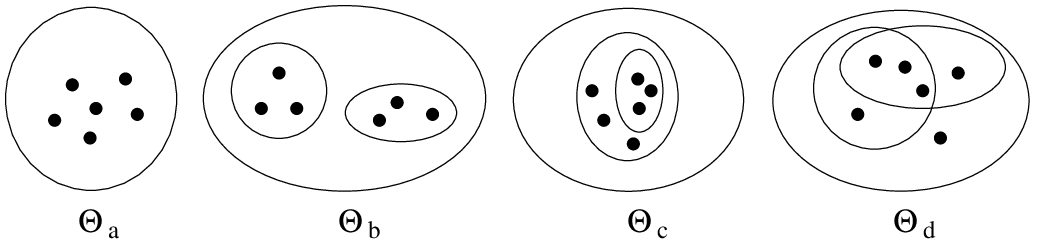}}}
\caption{Left, $\Theta_a$ entails contradictory possibilities represented as singletons. Centre-left,  $\Theta_b$ entails contradictory possibilities represented as sets that include further details. Centre-right, $\Theta_c$ represents two possibilities included in one another.  Right,  $\Theta_d$ represents two possibilities that have common elements but exhibit contradictions as well. This case has no counterpart in  PT.} \label{fig:Thetas}
\end{figure}

Let $\Theta$ denote a FoD. The left portion of Figure~\ref{fig:Thetas} illustrates the FoD as it appears in PT. In PT, the FoD contains only singletons --- for instance, the faces of a die. Thus, they can either be distinct or coincide, but cannot accomodate partial intersections. 

On the centre-left of Figure~\ref{fig:Thetas}, $\Theta_b$ is a FoD where two contradictory possibilities are represented as disjoint sets. They could represent two testimonies whose details point to different culprits, or the odd and even faces of a die.

On the centre-right of Figure~\ref{fig:Thetas}, $\Theta_c$ is a FoD where a second testimony shares certain details with the first one. Or, it could represent the even faces of a die as a subset of all faces.

On the right of   Figure~\ref{fig:Thetas} appear two possibilities that partially intersect. The two testimonies entail elements that reinforce one another, as well as elements that contradict one another. This case has no counterpart in PT.

The bulk of ET is concerned with procedures to combine partially contradictory, partially coherent testimonies. Testimonies, or \emph{bodies of evidence}, are assumed to arrive as sets of masses  supporting specific possibilities. 

Let us denote by $m(A_i)$ the mass of evidence supporting possibility $A_i$. Bodies of evidence take the form:

\begin{displaymath}
  \begin{array}{rcl}
    A & = & \{m(A_1), m(A_2), \ldots m(A_{N_A}) \} \\
    B & = & \{m(B_1), m(B_2), \ldots m(B_{N_B}) \} \\
     & \ldots & 
  \end{array}
\end{displaymath}

From Properties~(1) and~(2) follows:

\begin{enumerate}
  \item A positive mass $m(\Theta)>0$ can be assigned to the FoD. This mass hovers above the possibilities that are being envisaged, representing lack of information. In general, $m_A(\Theta) \neq$ $m_B(\Theta) \neq \ldots$.
  
  \item Since $\forall i,j$ it can be $A_i \cap A_j \neq \emptyset$, even if one adopts the normalization $\sum m(A_i) + m_A(\Theta) = 1$, in general $m(A_i \cup A_j) \neq$ $m(A_i) + m(A_j)$ 
\end{enumerate}
where similar considerations apply to any body of evidence $B$, $C$, $D$, etc.

With the caveats expressed in the above propositions, the following normalization is generally made:

\begin{equation}
  \sum_i m(A_i)  \; + \; m(\emptyset) \; + \; m(\Theta) \; = \; 1    \label{eq:normalization}
\end{equation}
where in general $m(\emptyset) =0$ except for a case that will be discussed in \S~\ref{subsec:ToM}.

Bodies of evidence must be interpreted by their recipient in terms of \emph{hypotheses} that they  formulates. Let us suppose that a recipient wants to evaluate to what extent a body of evidence $\{ m(A_1),$ $ m(A_2), \ldots$   $m(A_{N_A}),$  $m_A(\Theta) \}$  supports hypothesis  $\mathcal{H}$. The  \emph{Belief Function} expresses strict support for $\mathcal{H}$  by the available evidence:

\begin{equation}
  Bel(\mathcal{H}) \; = \;  \sum_{A_i \subset \mathcal{H}} \; m(A_i)    \label{eq:belief}
\end{equation}
where by definition $Bel(\Theta) = 1$ and  $Bel(\emptyset) = 0$.

By contrast, the  \emph{Plausibility Function} expresses partial support for $\mathcal{H}$:

\begin{equation}
  Pl(\mathcal{H}) \; = \;  \sum_{A_i \cap \mathcal{H} \neq \emptyset} \; m(A_i)   \label{eq:plausibility}
\end{equation}
where by definition $Pl(\Theta) = 1$ and  $Pl(\emptyset) = 0$.

In general, $Bel(\mathcal{H}) \leq Pl(\mathcal{H})$. Usage of belief or plausibility depends on applications.

The ensuing \S~\ref{sec:three} illustrates procedures for combining bodies of evidence that are appropriate for different classes of living beings. The combined evidence is then interpreted by means of eqs.~\ref{eq:belief} and~\ref{eq:plausibility}.

\section{Three Classes of Living Beings}    \label{sec:three}

Henceforth, I shall introduce rules to combine partially coherent, partially contradictory information by the following three classes of living beings:

\begin{enumerate}
\item Living organisms that react to sequences of stimuli which they are not able  anticipate.

\item Animals with a nervous system sufficiently sophisticated to anticipate future states.

\item Animals with a brain sufficiently sophisticated to figure out what others think about them.
\end{enumerate}

The first category certainly includes unicellular organisms, fungi, plants as well as animals with a purely reactive nervous system. 
One criterion to identify the threshold between (1) and (2) could be that of classifying in (2) those animal species that are capable of Pavlovian conditioning, whereas species in (1) do not.~\footnote{Ivan Pavlov experimented with dogs. Pavlov observed that dogs salivated when they a bell announced that they would be fed with meat, and that they salivated as soon as they heard the bell, even if no meat followed. Pavlovian conditioning requires the ability to anticipate the future (the meat) out of the present (the bell).} Research has continuously pushed down this threshold, first with identifying the capability of Pavlovian responses in specific fish species \cite{hollis-84JEP-ABP}, then extending this threshold to vertebrates \cite{fanselow-wassum-16CSHPB}, then finding a few simple invertebrates that are capable of Pavlovian learning albeit certain vertebrate species do not \cite{krause-domjan-17XII}.

The exact placing of this threshold is irrelevant to this paper, though it is key for it to recognise that one such threshold does exist. Notably, this marks a key difference with theories that take the ability to anticipate the future as a key definitory feature of life \cite{rosen-91} \cite{rosen-12}.

The  threshold between (2) and (3) is based on having a \emph{Theory of Mind} (ToM). ToM, also known as \emph{mentalizing}, \emph{meta-representation}, \emph{second-order intentionality} or \emph{mind-reading}, indicates the ability to figure out  what others think about oneself. It is a  sophisticated  ability that marks a sharp divide between humans and most  other animals, albeit certain primates appear to have it to some extent \cite{byrne-88tutto} \cite{byrne-95} \cite{devaine-sangalli-trapanese-bardino-hano-saintjalme-bouret-masi-daunizeau-17PLOS-CB}.  For instance, chimpanzees  are organized in  hierarchies headed by one male but, unlike most animals with similar social organizations, females and non-dominant males are capable of arranging  secret intercourses. Such arrangements, as well as those enacted in order to escape from the dominant male's wrath, point to the existence of a substantial degree of mind-reading \cite{byrne-95}.

The ability to think what others think can induce potentially infinite regressions on what possibilities  are being conceived, making social codes inherently unstable \cite{page-08JTP}. Thus, the  two codes whose existence has been recognized long before STE, namely the genetic code and human cultural communication codes, are extremes in terms of stability and instability, respectively.

The transitions between the above classes are not perfectly sharp, with a few species or specific individuals exhibiting the features of the superior class to some sort of intermediate extent. All what is needed is that the transition is sufficiently sharp to mark a qualitatively different way to handle information, which reflects into different versions of ET.

\begin{figure}
\center
\fbox{\resizebox*{0.8\textwidth}{!}{\includegraphics{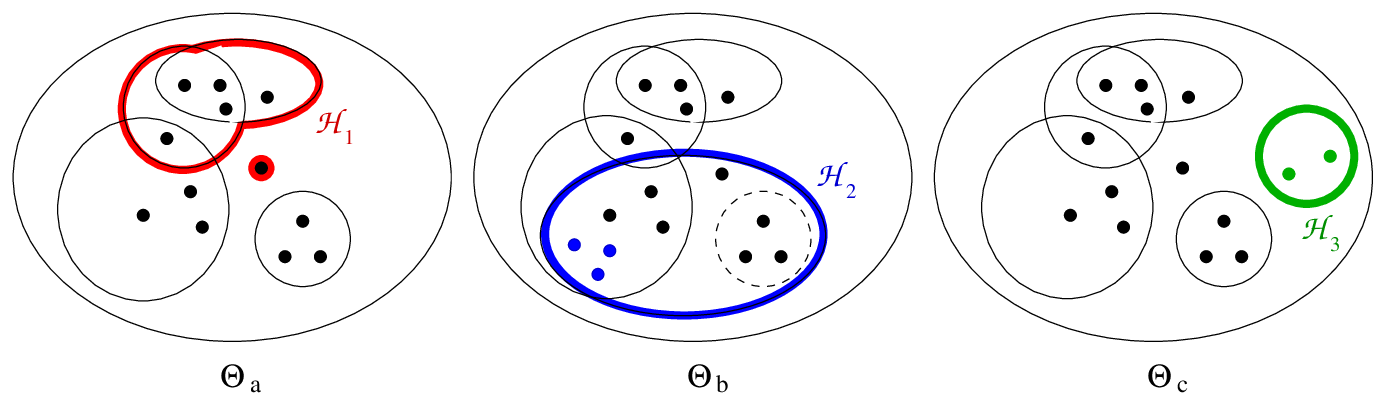}}}
\caption{Left, hypotesis  $\mathcal{H}_1$  is generated by combining existing possibilities. Centre, hypotyhesis $\mathcal{H}_2$ is  conceived by extending a previous possibility, now shown as a dotted circle. Right, hypothesis $\mathcal{H}_3$ explores unknown regions of the FoD in order to discover novel possibilities.} \label{fig:Hs}
\end{figure}

The three aforementioned capabilities  correspond to three ways of conceiving hypotheses. Figure~\ref{fig:Hs} shows on the left  a hypothesis $\mathcal{H}_1$  generated by combining possibilities that had been brought to the FoD by the bodies of evidence that it received. Centre, hypothesis $\mathcal{H}_2$ is conceived by extending a previously existing possibility (i.e., conceiving the possibility of smaller and more efficient microchips). On the right, novel possibilities are discovered by probing hitherto unexplored regions of the FoD with hypothesis $\mathcal{H}_3$. This is, in the legal-investigative framework of ET, the ability of Sherlock Holmes that Dr. Watson is unable to reach.

Finally, note that having  a sophisticated information processing capability does not imply that its corresponding capability is employed all the times. Individuals that are capable of (2) may resort to (1) when they face particularly simple tasks, or simply because of occasional failures. Likewise, being capable of (3)  excludes neither (2), nor (1). Human beings, as a matter of fact, do not always engage in difficult speculations of what all others think in their everyday lives, and at least sometimes  resort to instinct instead of extrapolating future states. What matters is that albeit all of these three modes imply interpreting information, their processes are quite different.

\subsection{Purely Reactive Organisms}   \label{subsec:genetic}

In this section I shall adapt ET to an organism receiving bodies of evidence that it is unable to compare to one another independently of arrival time. Nevertheless, this organism is capable of combining incoming bodies of evidence to a stored compound evidence somehow.

The simplest case is that of receiving organisms that ignore the time sequence of incoming bodies of evidence. Such is the case, for instance, of \emph{quorum sensing} employed by many insects and unicellular organisms \cite{hense-kuttler-muller-rothballer-hartmann-kreft-07NRM}.  However, since quorum sensing is  simpler than the case when the time sequence  matters, the general rule must apply to sequence sensing while eventually encompassing quorum sensing as a special case. 

For greater simplicity and without loss of generality let us consider a sequence of bodies of evidence $(A, B,\ldots)$ entailing only one possibility each (i.e., $A=\{A_1\}$, $B=\{B_1\}$, $C=\{C_1\}$  and so forth, where for simplicity pedices will be omitted henceforth).  Let us define a  \emph{Cumulative Combination Rule} as an algorithm composed by the following steps:

\begin{eqnarray}
  1. & Concatenation &  X_t \circ Y_{t+1} \; \mapsto \; XY_{t+1} = (X_t \cup Y_{t+1})  \nonumber \\
  2. & Sum \;  Evidence & m(XY_{t+1}) = m(X_t) + m(Y_{t+1})   \nonumber  \\
  3. & Normalization & if \;\; m(XY_{t+1}) \geq 1 \;\;\; m(XY_{t+1}):= 1 \nonumber \\
     & & m(\Theta):= 1 - m(XY_{t+1}) \;\; otherwise       \label{eq:cumulative}
\end{eqnarray}
$\forall X_t \in (A, AB, ABC \ldots)$ and $Y_{t+1} \in (A, B, \ldots)$. This algorithm is initialized with $X_0 = \emptyset$, $m(X_0)=0$ and $Y_1 = A$, $m(Y_1) = m(A)$.

\begin{figure}
\center
\fbox{\resizebox*{0.8\textwidth}{!}{\includegraphics{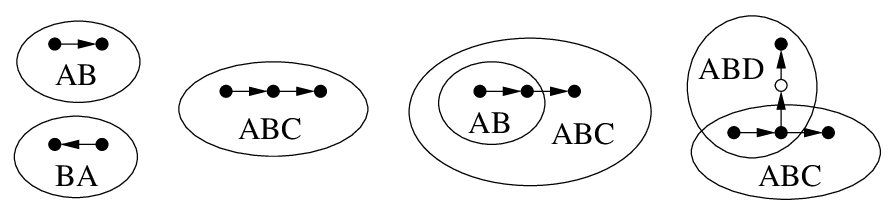}}}
\caption{Left, cumulative evidence $(AB)$ is different from cumulative evidence $(BA)$. Centre-left, the cumulative evidence $(ABC)$ conveyed by a codon. Centre-right, $(AB)$ obtained at $t=t_2$ is included in $(ABC)$ obtained at $t=t_3$. Right, if one nucleotide can be ignored (white circle), either $(ABC)$ or $(ABD)$ can be read.} \label{fig:abc}
\end{figure}

Rule~\ref{eq:cumulative}  distinguishes sequential ordering of evidence bodies but records their cumulative mass. For instance, as shown in the left portion of  Figure~\ref{fig:abc}, $AB$ is different from $BA$ although $m(AB) =$ $m(BA) =$ $m(A) + m(B)$. If this organism receives first $A$ and then $B$, it records $(AB)$ and $m(AB)$, whereas if it receives first $B$ and then $A$ it records $(BA)$ and $m(AB)$. However, it is unable to retrieve either $A$,  $m(A)$, or $B$, $m(B)$.

The special case of quorum sensing obtains when $Y_{t+1} \equiv X_t, \; \forall t$. In this case, recording a sequence $(AAA\ldots)$ is equivalent to recording $A$. Only the sum of masses of evidence matters, and only insofar it exceeds a threshold. 

Let us apply rule~\ref{eq:cumulative} to the interpretation of the genetic code by a cell synthetizing proteins. The genetic code is made of triplets of nucleotides (\emph{codons}) that map into amino acids that  subsequently join to form proteins. Within each codon, the sequence of nucleotides matters.

Let us consider the interpretation of one single codon. Let us represent its three nucleotides as bodies of evidence $A$, $B$, $C$ composed by one single element that arrive at $t=t_1, \: t_2, \: t_3$, respectively. A codon is a sequence $(ABC)$ as in the centre-left portion of Figure~\ref{fig:abc}. To fix ideas, let us assume that each nucleotide carries a mass of evidence equal to $1/3$ (other values could be assigned depending of physical-chemical constraints to receive specific nucleotides). By applying rule~\ref{eq:cumulative}, evidence is combined as follows:

\begin{displaymath}
  \begin{array}{rccrcl}
 t=t_1 & & &  A & = & \{m(A)= 1/3, \;\: m_A(\Theta)= 2/3 \} \\
 t=t_2 & & &  AB & = & \{m(A \: B)= 2/3, \;\: m_{AB}(\Theta) = 1/3 \} \\
 t=t_3 & & &  AB\,C & = & \{m(A \: B \: C)= 1 \} \\
  \end{array}
\end{displaymath}
where $m_A(\Theta)$ represents lack of information at $t=t_1$ whereas   $m_{AB}(\Theta)$ represents lack of information at $t=t_2$. At $t=t_3$, all information has finally become available.

Let us suppose that environmental feed-backs provide  tentative interpretations in terms of  different amino-acids, or hypotheses~$\mathcal{H}$. These hypotheses are of the simplest sort, assembled out of the   possibilities that are already envisaged in the FoD as illustrated in case (a) of Figure~\ref{fig:Hs}. Specifically, let us suppose that the sequence $(AB\,C)$ is recognized by amino-acid $\mathcal{H}_{ABC}$.

By applying eqs~\ref{eq:belief} and~\ref{eq:plausibility}, interpretation~$\mathcal{H}_{ABC}$ receives increasing support and is finally  confirmed:

\begin{displaymath}
  \begin{array}{rcl}
 t=t_1 & &  Bel(\mathcal{H}_{ABC}) = Pl(\mathcal{H}_{ABC}) = 1/3  \\
 t=t_2 & &  Bel(\mathcal{H}_{ABC}) = Pl(\mathcal{H}_{ABC}) = 2/3  \\
 t=t_3 & &  Bel(\mathcal{H}_{ABC}) = Pl(\mathcal{H}_{ABC}) = 1  \\
  \end{array}
\end{displaymath}

In reality, several codons can be interpreted as one and the same amino acid. Thus, the genetic code is a \emph{redundant}, or \emph{degenerate} code. It is not an ambiguous code, because the set of codons that correspond to a specific amino acid do not correspond to any other amino acid. Thus, the genetic code operates as in case (b) of Figure~\ref{fig:Thetas}. For instance, amino acid $\mathcal{H}_{ABC}$ could be produced by $(ZBC)$ as well, but both $(ABC)$ and $(ZBC)$ code for  $\mathcal{H}_{ABC}$ only.

The genetic code is  arbitrary  because, although (nearly) all existing organisms share the same code, experiments have shown that different correspondences between codons and amino acids are physically possible. Other correspondences might have existed in  the early stages of life. Moreover, the ancestral genetic code is likely to have been ambiguous, with one and the same codon corresponding to several amino acids \cite{woese-65PNAS}.

The simplest mechanism for the genetic code to have been ambiguous is that single nucleotides may have been misinterpreted, e.g., $(ABC)$ being misinterpreted as $(ABZ)$. However, other possibilities are more interesting.

Suppose, for instance, that it was not obvious for the ancestral code that the codons would be triplets. Suppose that either triplets or couples of nucleotides could be interpreted as amino acids. For instance, in the centre-right portion of Figure~\ref{fig:abc} one may either produce an amino acid as soon as $(AB)$ is received, or one may wait for $(ABC)$ to appear. Correspondingly, either amino acid $\mathcal{H}_{AB}$ or amino acid $\mathcal{H}_{ABC}$ is produced.

The sequences $(AB)$ and $(ABC)$ are included in one another as shown in the centre-right portion of Figure~\ref{fig:abc}. With the same (fictional) numbers as before, one obtains:

\begin{displaymath}
  \begin{array}{rclcl}
 t=t_1 & &  Bel(\mathcal{H}_{ABC}) = Pl(\mathcal{H}_{ABC}) = 1/3  & &  Bel(\mathcal{H}_{AB}) = Pl(\mathcal{H}_{AB}) = 1/3  \\
 t=t_2 & &  Bel(\mathcal{H}_{ABC}) = Pl(\mathcal{H}_{ABC}) = 2/3 &  & Bel(\mathcal{H}_{AB}) =  Pl(\mathcal{H}_{AB}) = 2/3 \\
 t=t_3 & &  Bel(\mathcal{H}_{ABC}) =  Pl(\mathcal{H}_{ABC}) = 1 &  &  Bel(\mathcal{H}_{AB}) = 2/3, \; Pl(\mathcal{H}_{AB}) = 1  \\
  \end{array}
\end{displaymath}

At $t=t_2$ these two interpretations have the same degrees of belief and plausibility. At $t=t_3$  $\mathcal{H}_{ABC}$ receives greater belief, but still the same plausibility as $\mathcal{H}_{AB}$. Thus, insofar these measures translate into interpreting the code as either $(AB)$ or $(ABC)$, this code is ambiguous.

One other possibility is that the ancestral code was made of triplets, but certain nucleotides could be occasionally ignored, in which case the first nucleotide of the subsequent codon would be interpreted as the last nucleotide of the current one. The right portion of Figure~\ref{fig:abc} illustrates this case, where either the codon $(ABC)$ or $(ABD)$ can be read.

Let us compute  belief and plausibility with the same numbers as before, except that $m(C) = $ $m(D) =$ $1/3 \cdot 1/2 = 1/6$. One obtains the same values for $(ABC)$ and $(ABD)$:

\begin{displaymath}
  \begin{array}{rclcl}
 t=t_1 & &  Bel(\mathcal{H}_{ABC}) = Pl(\mathcal{H}_{ABC}) = 1/3  & &  Bel(\mathcal{H}_{ABD}) = Pl(\mathcal{H}_{ABD}) = 1/3  \\
 t=t_2 & &  Bel(\mathcal{H}_{ABC}) = Pl(\mathcal{H}_{ABC}) = 2/3 &  & Bel(\mathcal{H}_{ABD}) =  Pl(\mathcal{H}_{ABD}) = 2/3 \\
 t=t_3 & &  Bel(\mathcal{H}_{ABC}) =  5/6, \;  Pl(\mathcal{H}_{ABC}) = 1 &  &  Bel(\mathcal{H}_{ABD}) = 5/6, \; Pl(\mathcal{H}_{ABD}) = 1  \\
  \end{array}
\end{displaymath}

This result is rather obvious, and could have been obtained by means of probabilistic resonings as well. What is interesting in this and the previous case is that whenever the code is ambiguous --- in the sense that several interpretations are possible --- there appears a discrepance between $Bel(\cdot)$ and $Pl(\cdot)$. In \S~\ref{sec:entropy}, this difference will be used to assess ambiguity reduction.

\subsection{Anticipatory Brains}          \label{subsec:wolf-snake}

ET  displays much wider potentialities once the capability of anticipating events is assumed. Hypotheses can be formulated, that go beyond the currently available evidence. Furthermore,  the coherence of available evidence can be evaluated independently of arrival time.

Let us assume that evidence   $A= \{m(A_1),$ $ m(A_2), \ldots$ $m(A_{N_A}),$ $m_A(\Theta)\}$ is available when a new body of evidence arrives,   $B = \{ m(B_1),$ $ m(B_2), \ldots$  $m(B_{N_B}),$  $ m_B(\Theta) \}$. Just like the sets entailed in one single body of evidence are not necessarily disjoint, $\forall i,j$ it may either be $A_i \subseteq B_j$, or $A_i \supseteq B_j$, or $Ai \cap B_j \neq \emptyset$, or $A_i \cap B_j = \emptyset$.

Dempster-Shafer's combination rule \cite{dempster-68JRSSB} \cite{shafer-76} yields the components of a new body of evidence $m_C$ that unites two bodies $m_A$ and $m_B$. Note that intersections with $\Theta$ enter the computation. 

\begin{equation}
  m(C_k) \; = \; \frac{\sum_{X_i \cap Y_j = C_k} \; m_A(X_i) \, m_B(Y_j)}{1 - \: \sum_{X_i \cap Y_j = \emptyset} \; m_A(X_i) \, m_B(Y_j)}  \label{eq:dempster-shafer}
\end{equation}
where $X_i \in \{ A_i \, \forall i, \; \Theta  \}$,  $Y_j \in \{ B_j \, \forall j, \; \Theta  \}$, and  where the $C_k$s are defined by all possible intersections of the $X_i$s with the $Y_j$s.

The numerator of eq.~\ref{eq:dempster-shafer} measures the extent to which the two bodies of evidence coherently support $C_k$, whereas the denominator measures the extent to which they are not contradictory with one another. Equivalently, one can say that the numerator expresses the logic of serial testimonies whereas the denominator expresses the logic of parallel testimonies \cite{shafer-86IJIS}.

Dempster-Shafer’s combination rule~\ref{eq:dempster-shafer} can be iterated to combine any number of evidence bodies. Its outcome is independent of the order in which evidence bodies are combined. In other words, the ability to memorize  and anticipate  allows to ignore the sequence of arrival of bodies of evidence in order to focus on their content.

Animals that are capable of anticipation can formulate possibilities that extend beyond the possibilities suggested by the available bodies of evidence, as illustrated in the central section of Figure~\ref{fig:Hs}. It is, essentially, the ability to extrapolate.

Let us fix ideas with the canonical example of a wolf chasing a prey that hides behind a tree. The wolf is able to extrapolate the prey's position in the next time unit, at least approximately. Let $A_1$ denote the area in its visual field where the wolf expects the prey to be in the next time unit. Suppose that the wolf assigns $m(A_1) = 0.8$ to its predictive capabilities, leaving a $m_A(\Theta) = 0.2$ for their failure.

Suppose that, after passing a tree, the prey disappears from sight. Let us denote this observation with $B_1$. Suppose that the wolf trusts its eyes and attention with $m(B_1) = 0.6$, leaving a $m_B(\Theta) = 0.4$ for mistakes. The two bodies of evidence are:

\begin{displaymath}
  \begin{array}{rclcl}
 A & = &  \{m(A_1) \; m_A(\Theta) \}  & = & \{0.8 \;\; 0.2 \}  \\
 B & = &  \{m(B_1) \; m_B(\Theta) \}  & = & \{0.6 \;\; 0.4 \}  \\
  \end{array}
\end{displaymath}

\begin{figure}
\center
\fbox{\resizebox*{0.8\textwidth}{!}{\includegraphics{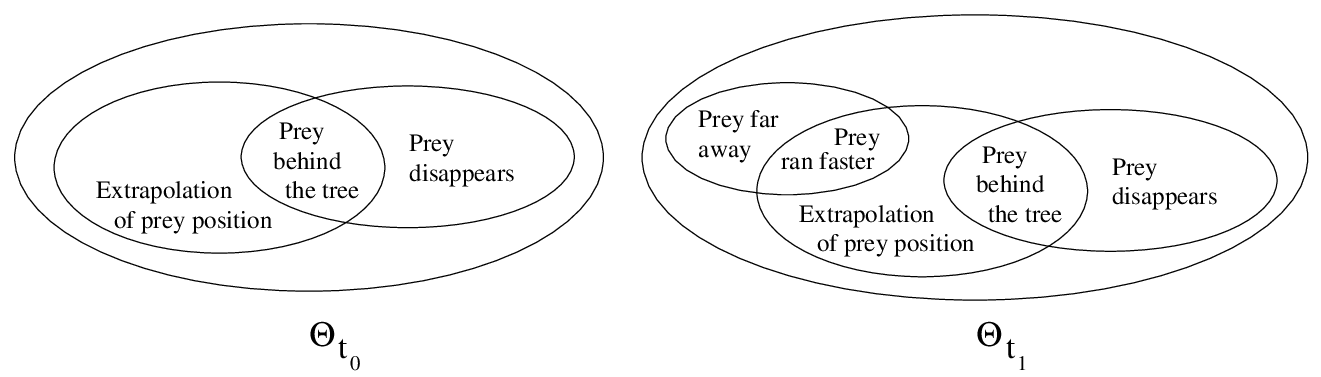}}}
\caption{Left, the initial FoD $\Theta_{t_0}$ with $A_1:$ \textit{Extrapolation of prey position}, $B_1:$ \textit{Prey disappears} and $C_1:$ \textit{Prey behind the tree}. Right, the FoD at a subsequent time step $\Theta_{t_1}$ to which $D_1:$ \textit{Prey far away} and $E_1:$ \textit{Prey ran faster} have been added.} \label{fig:wolf}
\end{figure}

Prey disappearance ($B_1$) is in contrast with the extrapolation of its position ($A_1$), but it is also possible that the prey is hiding behind the tree. Thus, $B_1$ and $A_1$ are not disjoint. The left portion of  Figure~\ref{fig:wolf} illustrates this FoD.

Let us denote their intersection with $C_1 = A_1 \cap B_1$. Let us rename $C_2 \equiv A_1$ and $C_3 \equiv B_1$. Eq.~\ref{eq:dempster-shafer} yields the combined body of evidence $C = \{m(C_1)$ $m(C_2)$ $m(C_3)$ $m_C(\Theta) \}$ whose masses are:

\begin{displaymath}
  \begin{array}{rclcl}
 m(C_1) & = &  m(A_1) \;m(B_1)   & = & 0.48  \\
 m(C_2) & = &  m(A_1) \;m_B(\Theta)  & = & 0.32  \\
 m(C_3) & = &  m_A(\Theta) \;m(B_1)  & = & 0.12  \\
 m_C(\Theta) & = &  m_A(\Theta) \; m_B(\Theta)   & = & 0.08  \\
  \end{array}
\end{displaymath}
which satisfy eq.~\ref{eq:normalization}.

The intersection  $C_1 = A_1 \cap B_1$ is very relevant for the wolf, because it means that the prey is hiding behind the tree. Thus, one may expect the wolf to formulate a hypothesis $\mathcal{H}_1 = C_1$, which receives the following support:

\begin{displaymath}
  \begin{array}{rclcl}
 Bel(\mathcal{H}_1) & = &  m(C_1)   & = & 0.48  \\
 Pl(\mathcal{H}_1) & = &  m(C_1) \; + \; m(C_2) \; + \; m(C_3)   & = & 0.92  \\
  \end{array}
\end{displaymath}

Now suppose that the wolf glimpses the prey further away. This is a new body of evidence, not quite reliable:

\begin{displaymath}
  \begin{array}{rclcl}
 D & = &  \{m(D_1) \; m_D(\Theta) \}  & = & \{0.2 \;\; 0.8 \}  \\
  \end{array}
\end{displaymath}

Body of evidence $D$ is still compatible with the wolf's extrapolations $A$, but it conflicts with the previous evidence about prey disappearance $B$. Thus, the possibilities appear now in the FoD as illustrated in the right portion of Figure~\ref{fig:wolf}.

Eq.~\ref{eq:dempster-shafer} can be used to combine $D$ with $C$ to obtain a new body of evidence $E$ where $E_1 \equiv C_1$,  $E_2 \equiv C_2$, $E_3 \equiv C_3$, $E_4 \equiv D_1$ and $E_5  = C_2 \cap D_1$ can be interpreted as the prey running faster than expected. Note that, since $D_1 \cap  C_1 = \emptyset$ and $D_1 \cap  C_3 = \emptyset$, the denominator of eq.~\ref{eq:dempster-shafer} is $1 - 0.2 \cdot 0.48 - 0.2 \cdot 0.12 =$ $0.88$. By applying  eq.~\ref{eq:dempster-shafer}  one obtains:

\begin{displaymath}
  \begin{array}{rclcl}
 m(E_1) & = &  m(C_1) \; m_D(\Theta) \; / \; (1 - m(D_1) \; m(C_3))   & \approx & 0.436  \\
 m(E_2) & = &  m(C_2) \; m_D(\Theta) \; / \; (1 - m(D_1) \; m(C_3))  & \approx & 0.291  \\
 m(E_3) & = &  m(C_3) \; m_D(\Theta) \; / \; (1 - m(D_1) \; m(C_3))  & \approx & 0.109  \\
 m(E_4) & = &  m_C(\Theta) \; m(D_1) \; / \; (1 - m(D_1) \; m(C_3))  & \approx & 0.018  \\
 m(E_5) & = &  m(C_2) \; m(D_1) \; / \; (1 - m(D_1) \; m(C_3))  & \approx & 0.073  \\
 m_E(\Theta) & = &  m_C(\Theta) \; m_D(\Theta) \; / \; (1 - m(D_1) \; m(C_3))   & \approx & 0.073  \\
  \end{array}
\end{displaymath}
which satisfy eq.~\ref{eq:normalization}.

The wolf is still interested in the hypothesis that the prey is hiding behind the tree, but
the intersection  $E_5 = C_2 \cap D_1$ is also relevant  because it means that the prey is still running. Let us assume that the wolf  formulates also a second hypothesis $\mathcal{H}_2 = E_5$. Let us compute belief and plausibility for these two hypotheses:

\begin{displaymath}
  \begin{array}{rclcl}
 Bel(\mathcal{H}_1) & = &  m(E_1)   & \approx & 0.436  \\
 Pl(\mathcal{H}_1) & = &  m(E_1) \; + \; m(E_2) \; + \; m(E_3)   & \approx & 0.836  \\
 Bel(\mathcal{H}_2) & = &  m(E_5)   & \approx & 0.073  \\
 Pl(\mathcal{H}_2) & = &  m(E_5) \; + \; m(E_4) \; + \; m(E_2)   & \approx & 0.382  \\
  \end{array}
\end{displaymath}

Hypothesis $\mathcal{H}_2$ receives substantially less support than $\mathcal{H}_1$, but support for $\mathcal{H}_1$ has decreased because conflicting evidence has arrived. In \S~\ref{subsec:ToM}, such a state of affairs eventually triggers the formulation of novel possibilities.

Dempster-Shafer combination rule~\ref{eq:dempster-shafer} works under the assumption that the decision-maker is capable of evaluating all pieces of evidence it has received hitherto, independently of arrival time. Differently from the cumulative combination rule~\ref{eq:cumulative} of \S~\ref{subsec:genetic}, the sequence of arrival does not matter.

In a way, one may maintain that one feature of memory  consists of enabling a brain to process information by evaluating exclusively its logical features (inclusions or intersections of possibilities). Spurious interpretations determined by arrival sequence --- e.g., sticking to the first interpretation simply because it was the first to appear --- can be overcome.

Similarly, extrapolation allows to anticipate  future states, inducing possibilities and formulating hypotheses that clearly derive from previously existing ones as shown in the central portion of Figure~\ref{fig:Hs}, but that are novel nonetheless. In a way, the information processors described in this section reduce both the past and the future to the present.

The canonical story about the wolf chasing a prey behind the tree continues with a snake who, contrary to the wolf, stops chasing its prey as soon as it hides behind the tree. No anticipation, mere reaction to a sequence of events as in \S~\ref{subsec:genetic}.

\subsection{Having a Theory of Mind}          \label{subsec:ToM}

One important consequence of having a ToM is that it generates indetermination of the possibilities that can be conceived. ToM can generate infinite regressions of the sort ``What is this person thinking about me?'', ``What is this person thinking that I am thinking about her?'', and so on. Even when thinking is restricted to one specific issue, ToM can slip into  regressions of the sort ``I think that you think that I think that...''

In practice, most of the times humans avoid  infinite regressions by limiting mind-reading to 2-3 levels \cite{basu-94AER} \cite{arad-rubinstein-12AER}, and in any case they appear to be incapable of more than 5 levels \cite{oesch-dunbar-17JN}. However, even limited levels of mind-reading can easily trigger the generation of a large number of possibilities  \cite{west-lebiere-01CSR}, marking a sharp transition of the number of hypotheses that humans and other primates can entertain.

Humans live in a social reality where novel possibilities continuously appear, and they are aware that they do. In contrast to probabilistic uncertainty, which concerns distributions of given possibilities, \emph{radical uncertainty} concerns  what possibilities  may appear in the FoD  \cite{runde-90EP} \cite{davidson-91JEP} \cite{dunn-01JPKE} \cite{dequech-04CJE} \cite{lane-maxfield-05JEE} \cite{kay-king-20}. Simple examples may include the uncertainty generated by novel technologies that may disrupt  current business plans as well as the effectiveness of specific weapons in warfare or, more in general, the uncertainty surrounding possible equilibria between world powers. Unlike probabilistic uncertainty, which can be hedged by proper insurance, radical uncertainty can have  dramatic consequences in terms of postponing or avoiding key decisions altogether \cite{tversky-shafir-92PSdeferred} \cite{yin-devreede-steele-devreede-19SS} \cite{yin-devreede-steele-devreede-21ICIS} \cite{gluth-rieskamp-buchel-13PLOSCB}. 

However, even if radical uncertainty is intractable by probabilistic methods, it does not eschew formalization altogether \cite{fioretti-25E}. Radical uncertainty originates from novel evidence that contradicts established causal relations, for the simple reason that once novel and unthinkable things have been observed, one  expects others to appear   \cite{locke-goldenbiddle-feldman-08OS} \cite{altmann-16II} \cite{saetre-vandeven-21AMR}.

In ET, two possibilities $A_i$ and $A_j$ conflict with one another if $A_i \cap A_j = \emptyset$. In standard ET, conflicting evidence is redistributed among available possibilities through the denominator of eq.~\ref{eq:dempster-shafer}.

By contrast, the Transferable Belief Model (TBM) assumes that conflicting evidence translates into $m(\emptyset) > 0$ \cite{smets-88} \cite{smets-92IJIS} \cite{smets-92XL}. The rationale of this assumption  is that conflicting evidence, by suggesting that something may happen, that is currently not imaginable, moves some mass $m$ towards the void set.

Correspondingly, the TBM substitutes Dempster-Shafer's with Smets' combination rule, which is essentially the numerator of eq.~\ref{eq:dempster-shafer}:

\begin{equation}
  m(C_k) \; = \; \sum_{X_i \cap Y_j = C_k} \; m_A(X_i) \, m_B(Y_j)  \label{eq:smets}
\end{equation}
where $X_i \in \{A_i \, \forall i, \: \emptyset, \: \Theta\}$,  $Y_j \in \{B_j \, \forall j, \: \emptyset, \: \Theta\}$ and the $C_k$s are defined by all possible intersections of the $X_i$s with the $Y_j$s.

Normalization is in order because eq.~\ref{eq:smets}, unlike eq.~\ref{eq:dempster-shafer}, does not combine bodies of evidence in ways that necessarily comply with eq.~\ref{eq:normalization}. In this respect, having a ToM  has similar consequences as the simple reception of a sequence of bodies of evidence analyzed in \S~\ref{subsec:genetic}.

The  Belief and Plausibility functions expressed by eqs.~\ref{eq:belief} and~\ref{eq:plausibility} must be amended on $\emptyset$ and $\Theta$  \cite{smets-07IF}. With eq.~\ref{eq:smets}, $Bel(\Theta) =$ $Pl(\Theta) = m_C(\Theta)$ and $Bel(\emptyset) =$ $Pl(\emptyset) = m_C(\emptyset)$.

Once conflicting evidence has been perceived, new possibilities must be explored and novel hypotheses must be formulated, as in the case depicted in the right portion of Figure~\ref{fig:Hs}. However, conflicting evidence expressed by $m(\emptyset)>0$ can trigger different reactions from different individuals, or from one and the same individuals at different points in time.

Consider classical detective stories. To be sure, Dr. Watson knows from the very beginning who's guilty. All clues point to one and only one culprit so if the case had been in his hands, Watson had simply closed it. However, Sherlock Holmes is profoundly disturbed by a tiny detail that contradicts the received interpretation. Thus, he interrogates other testimonies, finds other cues that do not fit with the rest of the picture, ascertains that certain testimonies are unreliable and, in the end, the denouement finally comes. Sherlock Holmes comes out with an entirely different interpretation, where certain details have a prominent place in causal explanations whereas others have been discarded.

ET understands the process of formulating novel hypotheses and looking for novel evidence, again and again until a coherent interpretation is reached, as refining and coarsening the FoD \cite{gordon-shortliffe-90VII-III} \cite{haenni-05WP} \cite{smets-07IF} \cite{shafer-16IJAR-hist}. This process is neither irrational nor obscure, but rather follows its own logic:

\bigskip

\begin{quotation}
Like any creative act, the act of constructing a frame of discernment  does not lend itself to thorough analysis. But we can pick out two considerations that influence it: (1) we want our evidence to interact in an interesting way, and (2) we do not want it to exhibit too much internal conflict.

Two items of evidence can always be said to interact, but they interact in an interesting way only if they jointly support a proposition more interesting than the propositions supported by either alone. (...) Since it depends on what we are interested in, any judgment as to whether our frame is successful in making our evidence interact in an interesting way is a subjective one. But since interesting interactions can always be destroyed by loosening relevant assumptions and thus enlarging our frame, it is clear that our desire for interesting interaction will incline us towards abridging or tightening our frame.

Our desire to avoid excessive internal conflict in our evidence will have precisely the opposite effect: it will incline us towards enlarging or loosening our frame. For internal conflict is itself a form of interaction --- the most extreme form of it. And it too tends to increase as the frame  is tightened, decrease as it is loosened.

\end{quotation}
Glenn 
Shafer \cite{shafer-76}, Ch. XII.

\bigskip

Interestingly, this quote entails a rationale for what Sherlock Holmes does --- tightening the FoD in order to highlight contradictions --- as well as what Dr. Watson does --- coarsening the FoD in order to arrive at a decision. Both directions must be pursued in order to enrich the FoD and formulate novel hypotheses. Note that the FoD is no longer a passive recipient of incoming bodies of evidence, because Sherlock Holmes actively looks for new evidence in order to resolve contradictions.

Note also that FoD tightening and coarsening occurs even in settings where mind-reading is unlikely to reach profound levels. Once evolution has endowed humans with ToM in order to manage complex societies \cite{dunbar-96}, humans apply it even in situations where simpler rules would suffice. Detective stories present us with contrived cases where the Sherlock Holmes are the heroes, but in everyday life, someone who treats any little issue the way Sherlock Holmes does is eventually ridiculed for being addicted to plot theories.

Just like animals that are capable of anticipation should not be expected to make use of it, humans may often find it more profitable not to exploit their ToM capabilities, and in many practical situations even the ability to extrapolate may be unnecessary. Humans often resort to simple heuristics that downgrade their behaviour to the purely reactive mode of unicellular organisms, but quite often this is sufficient, and certainly less costly \cite{gigerenzer-todd-99}.

\section{Evolutionary Pressures on Communication Codes}  \label{sec:entropy}

\begin{figure}
\center
\fbox{\resizebox*{0.8\textwidth}{!}{\includegraphics{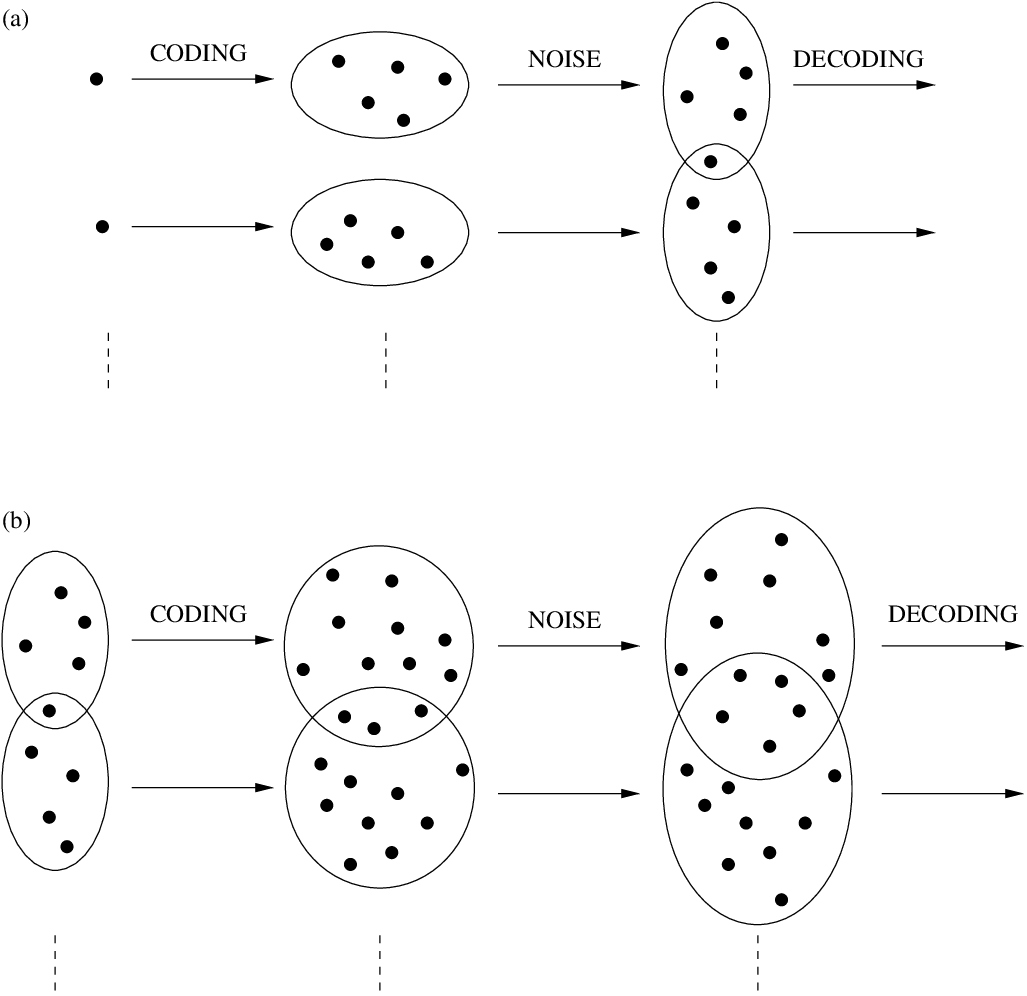}}}
\caption{On top (a),  classical IT where single characters $\{A_i\}$ are first coded into sets $A_i$, then transmitted through a noisy channel which may generate intersections between these sets, and finally decoded. Bottom (b), the fusion of partially overlapping information originating from different sources. The original overlap may be enhanced by coding and further enhanced by transmission through a noisy channel.} \label{fig:fusion}
\end{figure}

Similarly to the pairs of differences between ET and PT discussed in \S~\ref{subsec:ET}, ET can add the following features to IT:

\begin{enumerate}
  \item While Shannon's IT assumes that the source emits characters drawn from a given alphabet, known to the receiver, I shall henceforth assume that novel characters can be generated, unknown to the receiver. This is relevant for communication between living beings, because random mutations can generate novelties whose effects arbitrary communication codes can fan out.
  
  \item ET generalizes  IT with multiple sources emitting partially overlapping character sets (the evidence) whose overlap is further enhanced by coding characters into sequences and their transmission through a noisy channel. Figure~\ref{fig:fusion} illustrates this state of affairs.
\end{enumerate}

Adapting IT to the life sciences implies moving from an engineering setting where communication happens as planned, to an evolutionary system that is capable of generating unpredictable novelties that are subject to physical laws, nonetheless. In particular, life does not escape the general trend towards greater thermodynamic entropy, but   it is able to decrease macroscopic entropy --- the structures of living organisms --- by compensating it with higher entropy at more microscopic levels (e.g., heat dissipation) \cite{jeffery-pollack-rovelli-19E} \cite{jeffery-rovelli-20TN}.

Let us set aside the ncessary increase of microscopic entropy in order to focus on the structures of life as they can be grossly captured by information entropy. Since information entropy is greatest when all characters are equiprobable, increasingly structured living beings correspond to smaller values of information entropy in biological communication codes. Thus, let us take information entropy as a suitable Lyapunov function to be minimized by evolutionary systems (see \S~\ref{app:Lyapunov}).

Because of the above features~(1) and~(2), Shannon's information entropy requires some adaptation. The quest for a suitable entropy function  is a subject of debates that did not yet reach a universally accepted conclusion \cite{klir-06} \cite{abellan-17CSF} \cite{moralgarcia-abellan-21IJIS} \cite{dezert-tchamova-22IS} \cite{jirousek-kratochvil-shenoy-22IJAR}, but the following recent proposal \cite{ramisetty-jabez-subhrakanta-23IS}   is indicative of the sort of functionals that are being scrutinized:

\begin{equation}
  H \; = \; - \sum_{\mathcal{H}_i \in \Theta} \; \frac{Pl(\mathcal{H}_i) \;\; \lg \: Pl(\mathcal{H}_i)}{e^{Pl(\mathcal{H}_i) -Bel(\mathcal{H}_i)}} \;\; + \sum_{\mathcal{H}_i \in \Theta} \; \big[Pl(\mathcal{H}_i) -Bel(\mathcal{H}_i) \big]   \label{eq:panda}
\end{equation}

In eq.~\ref{eq:panda}, hypotheses $\mathcal{H}_i$ are formulated depending on the sort of capabilities illustrated in the introductory portion of \S~\ref{sec:three}, Figure~\ref{fig:Hs}. Hypotheses are formulated on combined evidence bodies obtained by means of the procedures illustrated in \S~\ref{subsec:genetic}, \S~\ref{subsec:wolf-snake} and~\ref{subsec:ToM}, respectively.

The first term of eq.~\ref{eq:panda}  reduces to Shannon's entropy if the $\mathcal{H}_i$s are singletons $\{A_i\}$s  and, consequently, $Bel(\mathcal{H}_i) =$ $ Pl(\mathcal{H}_i) =$ $ p(\{A_i\})$ where $p$ denotes probability. This term expresses contradiction of competing  evidence. The higher  this term, the more difficult an interpretation.

In the context of IT, this term can be minimized by adopting redundant codes that allow receivers to (partially) correct the mistakes introduced by the noisy channel (see the central portion of  Figure~\ref{fig:Thetas}). Living organisms do exploit this option; for instance, the genetic code is redundant (or \emph{degenerate}) and, while errors are most often made on the third nucleotide, this is precisely the one nucleotide on which most multiple codifications of one single amino acid differ from one another. However, one other option is available  to living organisms in order to minimize the first term of eq.~\ref{eq:panda}.

In IT, the receiver knows the alphabet of the source. Therefore, any character that has been received must belong to one of those in the alphabet. In IT, just like in PT, the set of possibilities is a given.

By contrast, living organisms can give novel meanings to novel  possibilities generated by either random mutations, or  random codings, or both. Whenever this happens, novel possibilities are added to the FoD, and by increasing the number of possibilities, information entropy can decrease \cite{atlan-74JTB} \cite{atlan-87PS}. This may have happened, for instance, each time the ancestral genetic code increased the number of amino acids that it codes from a likely initial number of about  10 to the current 20 amino acids.

The second term of eq.~\ref{eq:panda} has no counterpart in Shannon's entropy. The difference between $Pl(\mathcal{H}_i)$ and $Bel(\mathcal{H}_i)$ measures to what extent the available evidence goes beyond $\mathcal{H}_i$  to support  other hypotheses as well. Thus, it measures code ambiguity. Its minimization expresses the evolutiuonary trend towards less ambiguous codes; for  instance, the ambiguous ancestral genetic code has been substituted by the current non-ambiguous code.

To summarize, eq.~\ref{eq:panda} is a Lyapunov function whose minimization describes the evolutionary trends of organic communication codes  in terms of: (i) Reduction of communication errors; (ii) Appearance of novel meanings, and (iii) Reduction of ambiguity (see \S~\ref{app:Lyapunov} for details on Lyapunov functions).

\section{Conclusions}  \label{sec:conclusions}

STE moved its first steps in 1985 as a theory of life within theoretical biology \cite{barbieri-85}, but since at least 2001  it started to mutuate concepts from semiotics \cite{barbieri-24}. Albeit it seems natural for a research project based on meaning to pick concepts from the science of meanings, this transfer suffered from the fact that semiotics is tailored on humans, and while extensions to animals with a sufficiently developed brain are quite unproblematic, attributing sophisticated capabilities to unicellular organisms  either remains a purely formal and unnecessary renaming of existing concepts, or implicitly introduces a form of vitalism.

With this essay I am attempting to re-orient the field.  I made use of the physical theory of meaningful information \cite{rovelli-18III} \cite{kolchinsky-wolpert-18IF} as the basis of key concepts  such as ``meaning'' and ``interpretation.'' In this way, the higher-order concepts of semiotics can be still usefully applied to higher-order animals, but  are not fundamental to STE. STE can exist without semiotics, though it can make use of semiotics insofar it concerns cultural codes.

I introduced ET as a means to frame communication through arbitrary codes, but ET offers also an opportunity to revisit reasoning modes that are an integral part of semiotics. In particular, the creative mode illustrated \S~\ref{subsec:ToM} is nothing but the logic of abduction, precisely and restrictively defined as originating from cognitive conflicts that can only be resolved by either discovering details, or covering unnecessary ones, or both. Understanding abduction as triggered by cognitive conflicts and operating through iterative coarsenings and refinements of the FoD makes it a physically and psychologically grounded activity with specific features, rather than some sort of magical leap human minds are mysteriously capable of. If the presence of these features is required in order to speak of abduction, then certain purported instances of abduction rather appear as instances of induction, and \textit{vice versa}.

Similar considerations can be made concerning the ability to anticipate the future by extrapolating from the past and present considered in  \S~\ref{subsec:wolf-snake}. This ability requires nothing less than conceiving novel possibilities in a timeless space out of a few observed instances, which can be regarded as the psychological mechanism underlying induction. In settings certainly more complex than wolves hunting their preys, extrapolating from instances to timeless universals, be it enumerative or empirical  universals, and including universals corroborated by iterative rules as it is the case with mathematical induction \cite{rips-asmuth-07X}, means casting future, present and past possibilities as absolutes. Understanding induction as a sophisticated version of predicting a prey's position suggests a clear delimitation of this reasoning mode as characterized by smashing time-specific instances  into general, absolute and timeless possibilities.

Deduction is normally meant to be more complex than arranging sequences of stimuli into possibilities as illustrated in  \S~\ref{subsec:genetic}, but it has been suggested that causation originates from making choices --- for instance, the choices enabled by arbitrary codes --- in an environment where the trend towards greater thermodynamic entropy generates time irreversibility \cite{rovelli-21II} \cite{rovelli-23PP}. Thus, the most elementary processes considered in STE can be  viewed as generating causal relations, and therefore, deductions. In particular, establishing a causal derivation of specific amino acids from specific nucleotides could be regarded as the first, primary deduction that acted as a sort of template for the most complex  logical constructions that have been built  ever since.

Since making a choice implies generating information, the corresponding increase of thermodynamic entropy goes along with decreasing information entropy \cite{rovelli-24FP}. In \S~\ref{sec:entropy}, this is what happens to the ET-based extension of information entropy. One may also speculate that, within this process, the drive towards reducing ambiguity and the generation of novelties feed on one another, in the sense that novelties may generate ambiguities that are subsequently reduced.

Supported by the discovery of an increasing number of arbitrary codes, STE is pointing to possibilities for alternative sorts of life and its likely existence elsewhere in the universe. I hope to have shown that the framework of ET, based on communication and interpretation, is more appropriate than that of a gambler on a given possibility set.

\section*{Legal Disclaimers}

This research was not funded. The author has no conflict of interests. No copyrighted material was used without permission. Neither humans nor other animals were involved in experiments.

\appendix

\section{Lyapunov Functions}          \label{app:Lyapunov}

\begin{figure}
\center
\fbox{\resizebox*{0.7\textwidth}{!}{\includegraphics{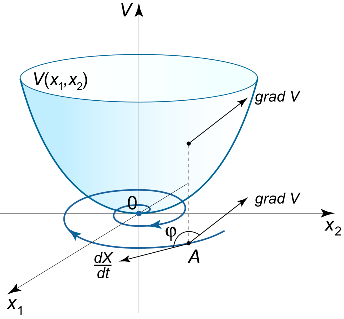}}}
\caption{A Lyapunov function $V$ of two state variables $x_1$ and $x_2$ describing movement towards equilibrium at $(0,0)$ along the path projected on the $x_1, x_2$ plane. A different function with a similar shape had worked equally well. By courtesy of \copyright Alex Svirin, $www.math24.net$.} \label{fig:lyapunov}
\end{figure}

A Lyapunov function can be used to prove the stability of an equilibrium point. A Lyapunov function is continuous, has continuous first derivatives, is strictly positive except for the equilibrium point, and its time derivative is non-positive. Figure~\ref{fig:lyapunov} illustrates a Lyapunov function for a system described by two state variables $x_1$ and $x_2$ with a stable equilibrium at $(0,0)$. The equilibrium is reached by minimizing $V$.

Several Lyapunov functions can exist for one and the same equilibrium point, all what is required is that the Lyapunov function has the required shape. For instance, the Lyapunov function of Figure~\ref{fig:lyapunov} would identify  $(0,0)$ as a stable equilibrium point even if the surrounding basing would be narrower, or wider than it is.

Lyapunov functions shaped like a Mexican hat can represent  the trend  towards a limit cycle between the edge of the hat and the height in the centre. More complex Lyapunov functions can entail several locally stable equilibria, in which case the Lyapunov function illustrates the capability to switch between different equilibria by jumping through saddles. Lyapunov functions cannot represent strange attractors.

The construction of a Lyapunov function is more an art than a science, though it is known that in simple cases with one equilibrium quadratic functions work. Construction is eased by the awareness that in general several Lyapunov functions can exist, and that any of them works.

Landscapes offer a simple and intuitive example of Lyapunov functions where rain drops move towards basins of attraction represented by lakes and, finally, the sea. Electric potential is a Lyapunov function for electrons moving towards the positive pole. For ecosystems, fitness is a Lyapunov function with a minus sign. In this case, $-V$ maximization takes the place of $V$ minimization.

\bibliographystyle{plain}
\bibliography{references}
\end{document}